# A Model-Driven Stack-Based Fully Convolutional Network for Pancreas Segmentation

Hao Li
*School of Biomedical Engineering, Shanghai Jiao Tong University*
Shanghai, China
imaginist@sjtu.edu.cn

Jun Li
*School of Biomedical Engineering, Shanghai Jiao Tong University*
Shanghai, China
dirk_li@sjtu.edu.cn

Xiaozhu Lin
*Department of Nuclear Medicine, Ruijin Hospital, Shanghai Jiao Tong University School of Medicine*
Shanghai, China
lxz11357@rjh.com.cn

Xiaohua Qian*
*School of Biomedical Engineering, Shanghai Jiao Tong University*
Shanghai, China
xiaohua.qian@sjtu.edu.cn

***Abstract*—The irregular geometry and high inter-slice variability in computerized tomography (CT) scans of the human pancreas make an accurate segmentation of this crucial organ a challenging task for existing data-driven deep learning methods. To address this problem, we present a novel model-driven stack-based fully convolutional network with a sliding window fusion algorithm for pancreas segmentation, termed MDS-Net. The MDS-Net's cost function includes a data approximation term and a prior knowledge regularization term combined with a stack scheme for capturing and fusing the two-dimensional (2D) and local three-dimensional (3D) context information. Specifically, 3D CT scans are divided into multiple stacks to capture the local spatial context feature. To highlight the importance of single slices, the inter-slice relationships in the stack data are also incorporated in the MDS-Net framework. For implementing this proposed model-driven method, we create a stack-based U-Net architecture and successfully derive its back-propagation procedure for end-to-end training. Furthermore, a sliding window fusion algorithm is utilized to improve the consistency of adjacent CT slices and intra-stack. Finally, extensive quantitative assessments on the NIH Pancreas-CT dataset demonstrated higher pancreatic segmentation accuracy and reliability of MDS-Net compared to other state-of-the-art methods.**

## I. Introduction

Accurate segmentation of the human pancreas from medical imaging data can be applied in many computer-assisted diagnosis and treatment systems for pancreatic cancer and other diseases. However, the pancreas segmentation problem still has several challenges. On the one hand, manual pancreas labeling from magnetic resonance imaging (MRI) or computerized tomography (CT) data is costly, time-consuming. On the other hand, it is very challenging to obtain consistent features from the data of different patients through learning approaches due to the small data size and variability of the pancreas shape. As show in Fig. 1, these variations attest to the significant differences in the pancreas morphology at different CT slices. Traditional image segmentation algorithms achieved low Dice coefficients (<75%) on pancreas segmentation. Thus, new algorithms are needed to overcome the challenges of pancreas segmentation, and thereby achieve satisfactory segmentation performance.

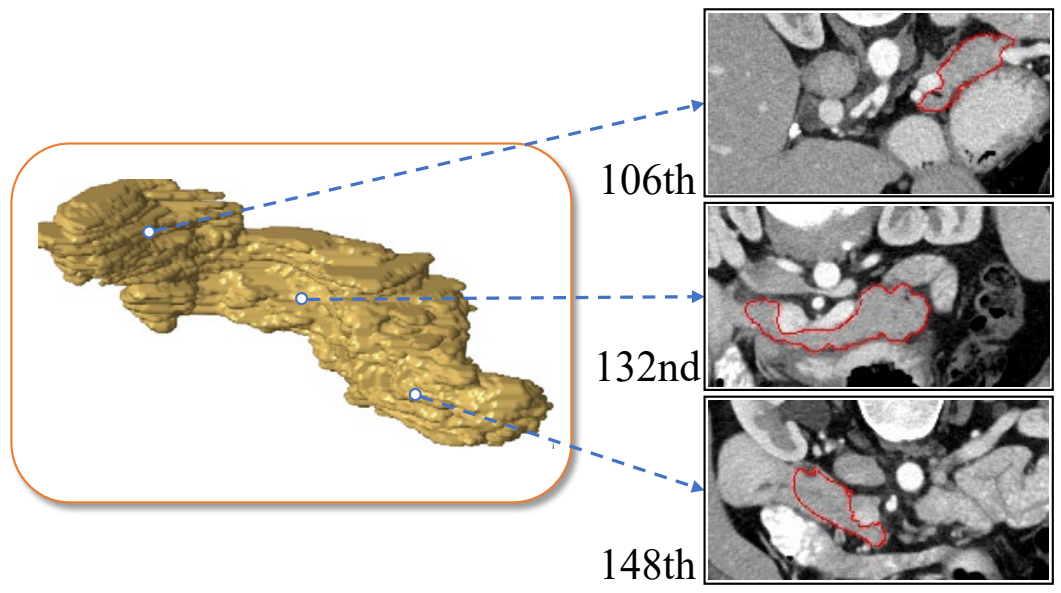

Fig. 1. Illustration of significant differences of pancreas morphology in different CT slices.

Nowadays, deep convolutional neural networks (CNNs) have demonstrated great potentials for medical image processing [1]. Segmentation approaches based on CNNs can be mainly divided into two types: two-dimensional (2D) networks and three-dimensional (3D) networks. The advantages of 2D segmentation networks, like FCN [2] and U-Net [3], have been well demonstrated in medical image processing [4]. Roth *et al.* [5] proposed a CNN-based holistically-nested network (HNN), which integrates semantic mid-level cues of deeply-learned interior and boundary maps of the target organ, finally achieved an average Dice coefficient of 78%. Zhou *et al.* [6] proposed a fixed-point FCN model for pancreas segmentation in abdominal CT scans, they obtained 82.4% Dice coefficient on NIH dataset. However, the pancreatic shape, size, and geometry in different CT slices are quite different, as shown in Fig. 1. Pancreas segmentation methods based on 2D slices ignore the pancreas continuity in the three-dimensional (3D) space, limiting further improvements in the segmentation performance. 3D CNN segmentation architectures, including V-Net [7] and 3D U-Net [8], can directly extract features from 3D spatial information, thus avoiding the bottleneck problem of a 2D segmentation network. However, due to the large GPU memory requirements of 3D networks, the input 3D CT scans are usually cropped to small chunks or down sampling to small volumes, which limits the spatial context learning.

Deep learning networks are mainly data-driven models whose performance is generally restricted by the scale of datasets. A primary reasons is that most of the existing deep learning models are black-box models and lack theoretical understanding and interpretability of the relationship between the network topology and performance [9]. Besides, the complexity of network design is also a limitation of data-driven deep learning methods, which restricts their generalization performance [10]. The abovementioned limitations can be remedied by model-driven deep learning approaches which effectively combine model-driven and data-driven methods through introducing prior knowledge and theoretical interpretations into deep learning frameworks. Herein, a model-driven deep learning approach refers to a method of constructing a model (e.g., a loss function) based on a specific task-based target, a physical mechanism, or domain knowledge. Recently, Lin *et al.* [11] presented an integration of model- and data-driven methods for synchronous adaptive multi-band image fusion. This integration resulted in image fusion outcomes of higher contrast, better visual perception, and less distortion. Zhang *et al.* [12] proposed combining data-driven and model-driven methods for robust facial landmark detection. These examples collectively demonstrate the effectiveness and increasing applications of model-driven approaches.

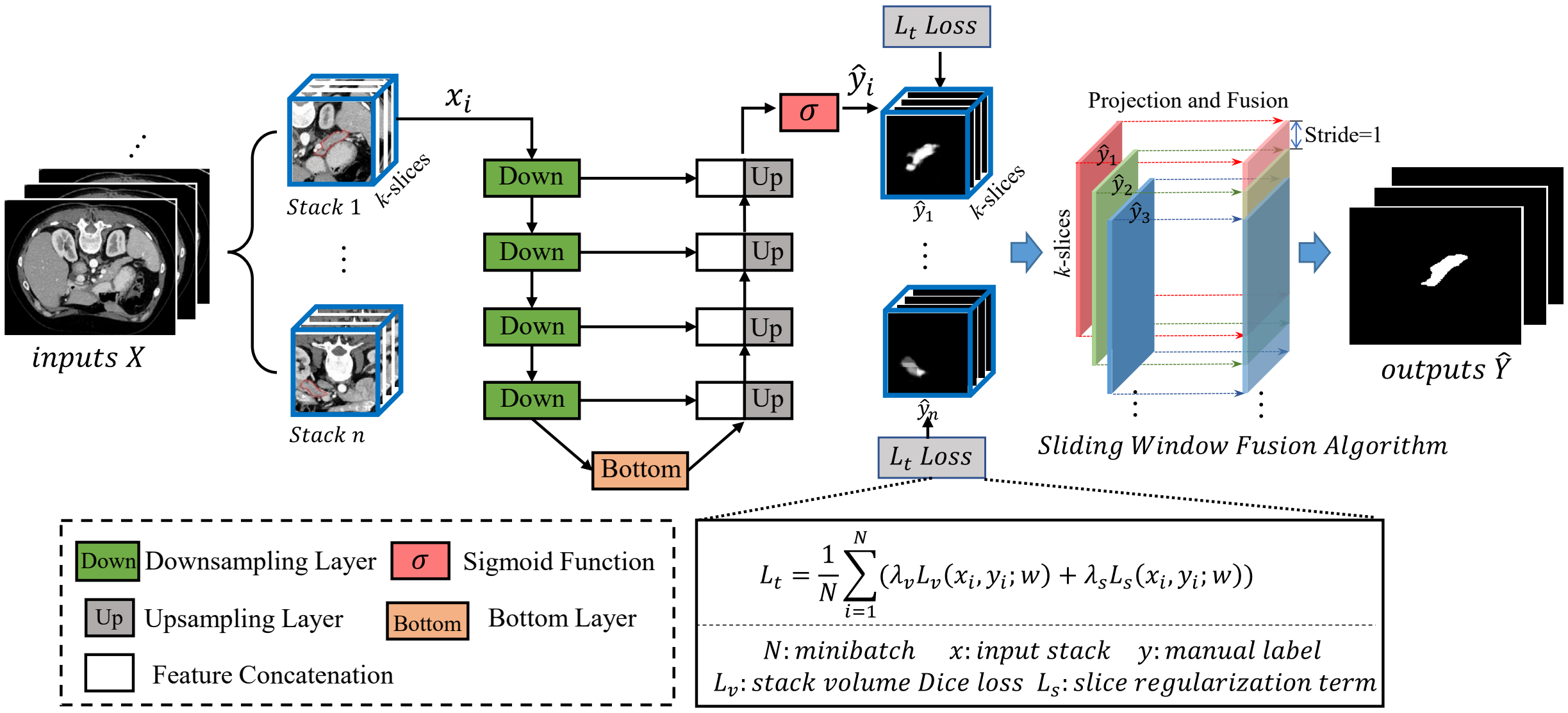

Fig. 2. The overall architecture of the MDS-Net model.

In this work, we use a model-driven deep learning method to introduce the spatial pancreas structure as a prior knowledge into the deep learning framework for pancreas segmentation. Precisely, the key contributions of the proposed MDS-Net framework can be summarized as: i) A stack-based U-Net architecture was developed to capture the local spatial context feature; ii) a model-driven regularization term was embedded into the cost function to integrate inter-slice relationship and prior knowledge constraints; iii) A sliding window fusion algorithm was used to refine the intra-stack and inter-slice consistency of the segmentation results; iv) The MDS-Net provides a potential method for those challenging segmentation tasks of small and complex human tissues, vessels or organs like the pancreas.

## II. METHOD

The overall framework of our proposed MDS-Net model is shown in Fig. 2. Assume that we are given 3D CT scans $X \in R^{d\times h\times w}$, where *d, h, w* are the depth, height, and width of the 3D CT are scans, respectively. The ground-truth $Y \in R^{d\times h\times w}$ and the predicted ones $\hat{Y} \in R^{d\times h\times w}$ have the same size as *X*. First, we extract *k*-slice stacks data $\{x_1, x_2, \dots, x_n\} \in R^{k\times h\times w}$ from 3D CT scans by sliding window algorithm, the window size is *k* and the slide stride is one. Then, an encoder-decoder model-driven Stack-U-Net (i.e., MDS-Net) model is utilized to obtain the predicted probability volumes $\hat{y}_i$. The MDS-Net energy function accounts for the overall stack loss and the inter-slice regularization term. This energy function can capture and fuse local 3D spatial information and 2D contextual information, thereby ensures the accuracy of both the single-slice and overall segmentation results. Finally, sliding window fusion algorithm is used to refine the voxel-wise segmentation results. In the following subsections, we will introduce: 1) the MDS-Net model structure and design; 2) back-propagation rule derivation; 3) sliding window fusion algorithm.

### A. MDS-Net Model Structure and Design

Two common segmentation metrics, namely the Dice coefficient ($Dice\ loss = 1 - 2(|Y \cap \hat{Y}|)/(|Y| + |\hat{Y}|\ )$) and the Jaccard index ($Jacccard\ loss = 1 - |Y \cap \hat{Y}|/|Y \cup \hat{Y}|\ $), are generally used as cost functions in deep-learning-based segmentation methods [3, 13]. Here, based on the Dice loss function, we proposed a new cost function that accounts for local spatial information of stack data into the segmentation process. The cost function can be formulated as,

$$L_v(x, y; \theta) = 1 - \frac{2|y \cap f(x;\theta)|}{(|y| + |f(x;\theta)|)} \tag{1}$$

Where *f* is the segmentation model, *x* denotes the stack data input, $\theta$ is the network parameters, and *y* is the corresponding manual label. According to the input *x* and network parameters w, the local stack prediction result $\hat{y}$ can be obtained by *f*. The new cost function $L_v$ is used to measure the overlapping rate of *y* and $\hat{y}$ at stack level. This function is designed based on a divide-and-conquer scheme, where a whole set of 3D CT scans $X \in R^{d\times h\times w}$ is divided into multiple 3D $k$-slice stacks $\{x_1, x_2, \dots, x_n\} \in R^{k\times h\times w}$ , $k < d$ . Network training is conducted with the 3D stack units, whose results are finally integrated into the overall pancreas segmentation results.

While the local spatial information is accounted for in the $L_v$ (1), single-slice segmentation cost is overlooked. Thus, we added a regularization term to the $L_v$ cost function to control the relationship between the overall slice information and the single-slice information. The addition of this term improves the single-slice segmentation quality, the regularization term is formulated as,

$$L_s(x, y; \theta) = \left(\sum_{m=1}^{k}\left(1 - 2\frac{|y^m \cap f^m(x;\theta)|}{|y^m| + |f^m(x;\theta)|}\right)^2\right)^{\frac{1}{2}} \tag{2}$$

Where $y = \{y^1, y^2, \dots, y^k\}$, $y^m$ is the $m$-th slice in stack manual label $y$, and $f^m(x; w)$ is the corresponding prediction result of $m$-th slice, $f^m(x; w)$ can also be denoted as $\hat{y}^m$. Thus, the proposed cost function consists of two parts: the first part, $L_v$, measures the approximation loss of the 3D stack data, while the second part, $L_s$, regularizes the loss of each CT slice. The overall energy function can be formulated as,

$$L_t = \frac{1}{N}\sum_{i=1}^{N}\left(\lambda_v L_v(x_i, y_i; \theta) + \lambda_s L_s(x_i, y_i; \theta)\right) \tag{3}$$

where $N$ is the batch size of the segmentation model, $x_i$ is the $i$-th stack data in a mini-batch, $y_i$ is the corresponding manual label of $x_i$ . $\lambda_v$ and $\lambda_s$ are the coefficients of the approximation term $L_v$ and the regularization term $L_s$.

Based on the overall loss defined in (3), we proposed MDS-Net, which represents an encoder-decoder stack-based U-Net architecture. 3D stack data $x \in R^{k\times h\times w}$ is fed into MDS-Net in a multi-channel manner to learn local spatial context information. For constructing segmentation network, we follow the classic encoder-decoder ResUNet architecture, which consists of downsampling and upsampling phases. The downsampling unit is composed of basic residual module and max-pooling layers, and the upsampling unit is composed of de-convolutional layer and residual module. Fig 3 illustrates the detailed architecture of the residual module.

The pancreatic structure is strongly correlated in adjacent slices while a weak association exists between slices that are distant from the abdominal CT scan. Thus, it is sufficient for MDS-Net to obtain the local spatial features from stacks data (i.e., the upper and lower CT slices), in order to improve the pancreas segmentation accuracy. Moreover, the MDS-Net parameters are much fewer than those of other 3D segmentation networks. Hence, MDS-Net has lower computational complexity to a large extent, improved training efficiency, and higher network portability without compromising the segmentation performance.

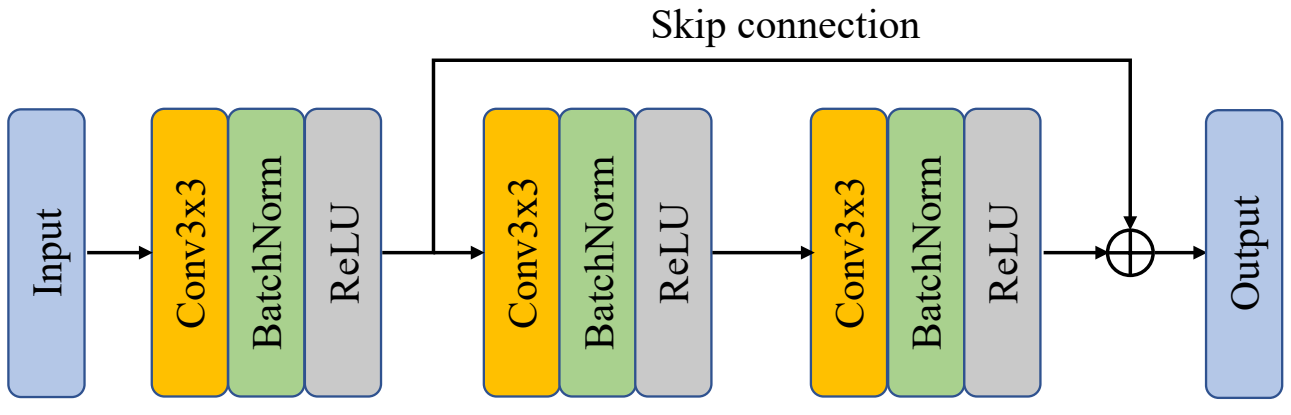

Fig. 3. Basic residual module in MDS-Net.

*B. Back-propagation for the MDS-Net Energy Function*

The components of the MDS-Net energy function are all differentiable. Therefore, it is relatively straightforward to obtain the derivatives constituting the gradients required by the back-propagation (BP) algorithm. Since the Dice loss function is the key term of $L_t$, the derivation process can be simplified by setting,

$$\delta_{\hat{Y}} = \partial Dice(\hat{Y}, Y)/\partial\hat{Y} \tag{4}$$

where $\delta_{\hat{Y}}$ denotes the partial derivative of the Dice loss with respect to the prediction result $\hat{Y}$ . The Dice loss itself is differentiable, and this fact has been applied in many segmentation problems. Similarly, to simplify derivations, we neglect the mini-batch influence on the loss. So, the simplified loss $L_t$ becomes,

$$\begin{aligned} L_t &= \lambda_v L_v(x, y; w) + \lambda_s L_s(x, y; w) \\ &= \lambda_v\left(1 - \frac{2|y \cap f(x;w)|}{|y| + |f(x;w)|}\right) + \lambda_s\left(\sum_{m=1}^{k}\left(1 - \frac{2|y^m \cap f^m(x;w)|}{|y^m| + |f^m(x;w)|}\right)^2\right)^{\frac{1}{2}} \qquad (5) \\ &= \lambda_v Dice(\hat{y}, y) + \lambda_s\|Dice(\hat{y}^m, y^m)\|_2 \end{aligned}$$

where $\hat{y} = f(x; w)$, $\hat{y}^m = f^m(x; w)$ denote the prediction results at stack and slice level, and $\|\cdot\|_2$ is the L2 norm operation. For the prediction result of $p$-th slice $\hat{y}^p$, the partial derivative of $L_t$ with respect to $\hat{y}^p$ is derived follows,

$$\begin{aligned} \frac{\partial L_t}{\partial\hat{y}^p} &= \frac{\partial L_t}{\partial L_v(x,y;w)}\frac{\partial L_v(x,y;w)}{\partial\hat{y}}\frac{\partial\hat{y}}{\partial\hat{y}^p} + \frac{\partial L_t}{\partial L_s(x,y;w)}\frac{\partial L_s(x,y;w)}{\partial\hat{y}^p} \\ &= \lambda_v\frac{\partial Dice(\hat{y},y)}{\partial\hat{y}} + \lambda_s\frac{\partial Dice(\hat{y}^p,y^p)}{\left(\sum_{m=1}^{k} Dice(\hat{y}^m,y^m)^2\right)^{\frac{1}{2}}}\frac{\partial Dice(\hat{y}^p,y^p)}{\partial\hat{y}^p} \\ &= \lambda_v\delta_{\hat{y}} + \lambda_v\delta_{\hat{y}^p}\left(\sum_{m=1}^{k} Dice(\hat{y}^m, y^m)^2\right)^{-\frac{1}{2}}Dice(\hat{y}^p, y^p) \qquad (6) \\ &= \lambda_v\delta_{\hat{y}} + \lambda_v\delta_{\hat{y}^p}\frac{Dice(\hat{y}^p, y^p)}{\|Dice(\hat{y}^m, y^m)\|_2} \end{aligned}$$

After calculating the derivatives of the MDS-Net energy function, the rest of the back-propagation process is the same as that for conventional CNNs. This formulation shows that the gradient for each slice is decided by the slice itself and by the information from the overall stack data. Hence, our proposed model can effectively integrate local spatial information of the stack data and ensure high segmentation accuracy for each single slice.

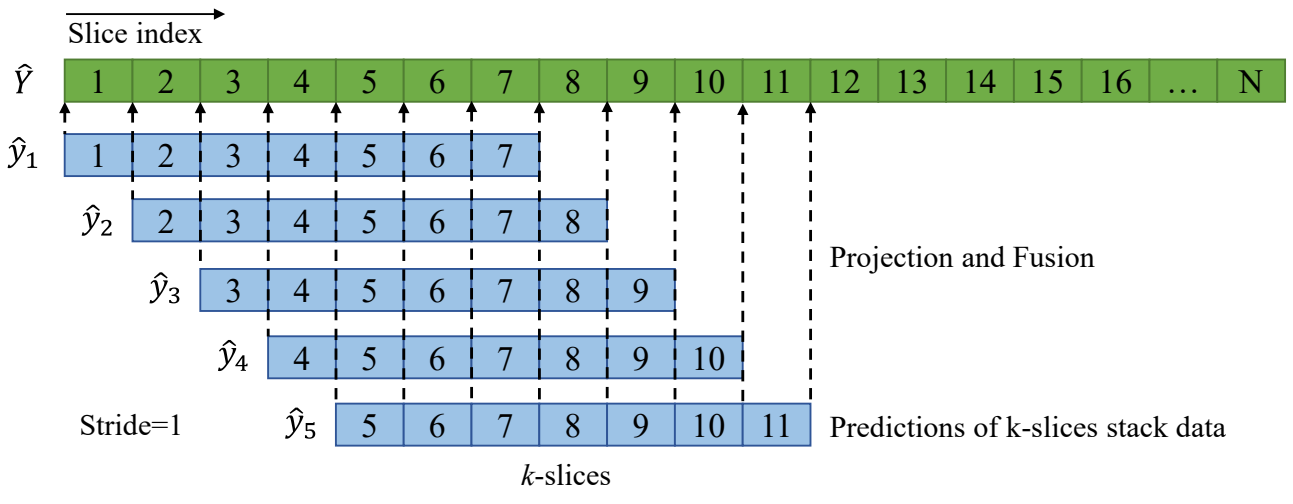

Fig. 4. The projection and embedding process of the sliding window fusion algorithm.

### C. Sliding Window Fusion Algorithm

Although the proposed model in (3) can capture the local spatial context information from the stack, there may exist subtle discrepancy between the adjacent stacks. Hence, based on MDS-Net, we introduced a sliding window fusion algorithm to enhance the inter-slice continuity of segmentation results.

Fig 4 illustrates the projection and embedding process of the sliding window fusion algorithm. Given a 3D CT scans input $X \in R^{d\times h\times w}$, we first extract a stack data sequence from $X$ by sliding window algorithm with $k$ windows size and one stride. The input stack data sequence can be denoted as $\{x_1, x_2, \ldots, x_{d-k+1}\} \in R^{k\times h\times w}$; adjacent stacks' overlap is *k-1* slices. Then, we can get a corresponding probability map sequence $\{\hat{y}_1, \hat{y}_2, \ldots, \hat{y}_{d-k+1}\} \in R^{k\times h\times w}$ predicted by MDS-Net. Finally, each $k$-slices stack prediction is projected and fused to its original position, and the final segmentation result $\hat{Y}$ is calculated by thresholding the average of the probability maps.

Although the overall shape of the pancreas is irregular, there is a strong relationship between the upper and lower layers of the local space. Therefore, the sliding window fusion algorithm can effectively ensure the smoothness of the segmentation result between adjacent stacks and slices, and further improve the pancreas segmentation performance.

## III. Experimental Results

### A. Dataset and Evaluation Criteria

Our proposed MDS-Net framework was evaluated on the NIH-CT dataset [14], developed by the US National Institute of Health (NIH), which contains 82 abdominal enhanced 3D CT scans. In the direction of the axial viewpoint, the CT slice size is 512*512 pixels, and the number of slices varies from 181 to 466 for different patients.

Experimental results were validated by 4-fold cross-validation (CV). To alleviate the overfitting problem, we augmented the training data set via rotation, horizontal flipping, and vertical flipping. The Dice similarity coefficient (DSC), the Jaccard index, the pixels-wise precision, and recall were used to evaluate our segmentation results. In addition, the root-mean-squared error (RMSE) [15] was utilized to measure the Euclidean distance between the edge contour of the segmentation result and the manual label contour. The formula of RMSE is defined in Eq. 11, where $(\hat{x}_i, \hat{y}_i)$ is the pixel coordination on the boundary of prediction result, and $(x_i, y_i)$ is the corresponding pixel coordination on label's boundary. Above evaluation criteria are formulated as follows,

$$DSC(A,B) = \frac{2\times|A\cap B|}{|A|+|B|} \tag{7}$$

$$Jaccard(A,B) = \frac{|A\cap B|}{|A\cup B|} \tag{8}$$

$$Precision = \frac{TP}{TP+FP} \tag{9}$$

$$Recall = \frac{TP}{TP+FN} \tag{10}$$

$$RMSE = \left(\frac{1}{n}\sum_{i=1}^{n}(\hat{x}_i - x_i)^2 + (\hat{y}_i - y_i)^2\right)^{\frac{1}{2}} \tag{11}$$

### B. Implementation Details

Our proposed network framework was implemented on PyTorch with NVIDIA GeForce GTX 1080Ti (11GB memory). Unless otherwise specified, each stack contained 7 CT slices, $\lambda_v$ and $\lambda_s$ were both set to 0.5 and the epoch was set to 10. We set the learning rate to *1e-4*, the batch size to 8 and the momentum value in SGD to 0.99. In addition, the MDS-Net network model was built on the coronal, sagittal and axial views, respectively, and their segmentation results were then fused to obtain an overall result. The average time of model training was approximate ~12 hours for MDS-Net on a single standard NVIDIA GeForce GTX 1080Ti (11GB memory)

### C. Qualitative and Quantitative Analysis

The quantitative 4-fold CV was implemented to evaluate the effects of our proposed MDS-Net model in terms of the DSC, Jaccard, Precision, Recall and RMSE on the HIN-CT dataset, as shown in Table I. The average accuracy of DSC, Jaccard, Precision and Recall on the 82 cases is 85.7%, 75.3%, 87.4%, and 84.8% respectively, demonstrating the excellent performance of MDS-Net on pancreas segmentation. The mean RMSE is 2.8 mm that indicated good conformity of predicted results to the reference boundaries. In addition, the small standard deviation and subtle fluctuation among different fold's results reflect the robustness and reliability of our proposed method.

Fig. 5 provides the 2D visual results of three slices with the U-Net, MDS-Net, and ground truth from the pancreatic head, middle, and tail parts, respectively. We can see that our proposed MDS-Net obtained a better performance on pancreas segmentation than U-Net, and effectively alleviate many error phenomena like over-segmentation, under-segmentation and misalignment. The pancreatic head and tail parts in CT scans are hard to segment accurately, due to the small area, variable shape, and low contrast of these regions, as shown in Fig. 5(a) and Fig5(c). Even in those cases, our MDS-Net can still achieve an ideal segmentation result, reflecting the outstanding ability of MDS-Net for small-target segmentation.

TABLE I. QUANTITATIVE EVALUATION OF THE SEGMENTATION RESULTS

| | mean ± stdv [min, max] | | | | |
|---|---|---|---|---|---|
| | DSC (%) | Jaccard (%) | Precision (%) | Recall (%) | RMSE (mm) |
| fold 0 | 85.4±4.3[75.7,90.6] | 74.8±6.4[60.9,82.8] | 85.4±6.9[66.6,94.3] | 86.3±7.4[60.9,93.8] | 3.2±1.4[0.7,6.1] |
| fold 1 | 86.7±3.9[73.6,91.3] | 76.7±5.8[58.3,84.0] | 87.7±2.8[80.6,91.7] | 86.1±6.9[64.1,92.7] | 2.5±1.1[1.4,6.1] |
| fold 2 | 84.7±4.6[73.2,91.6] | 73.8±6.7[57.7,84.5] | 88.1±5.1[77.5,95.7] | 82.3±8.0[64.0,93.0] | 2.6±0.7[1.6,4.0] |
| fold 3 | 86.0±3.4[78.1,90.9] | 75.7±5.1[64.1,83.3] | 88.3±4.5[80.4,97.8] | 84.5±7.2[90.8,65.0] | 2.9±1.4[1.4,6.8] |
| **overall** | **85.7±4.1[73.2,91.6]** | **75.3±6.1[57.7,84.5]** | **87.4±5.2[66.6,97.8]** | **84.8±7.5[64.0,93.8]** | **2.8±1.2[1.4,6.8]** |

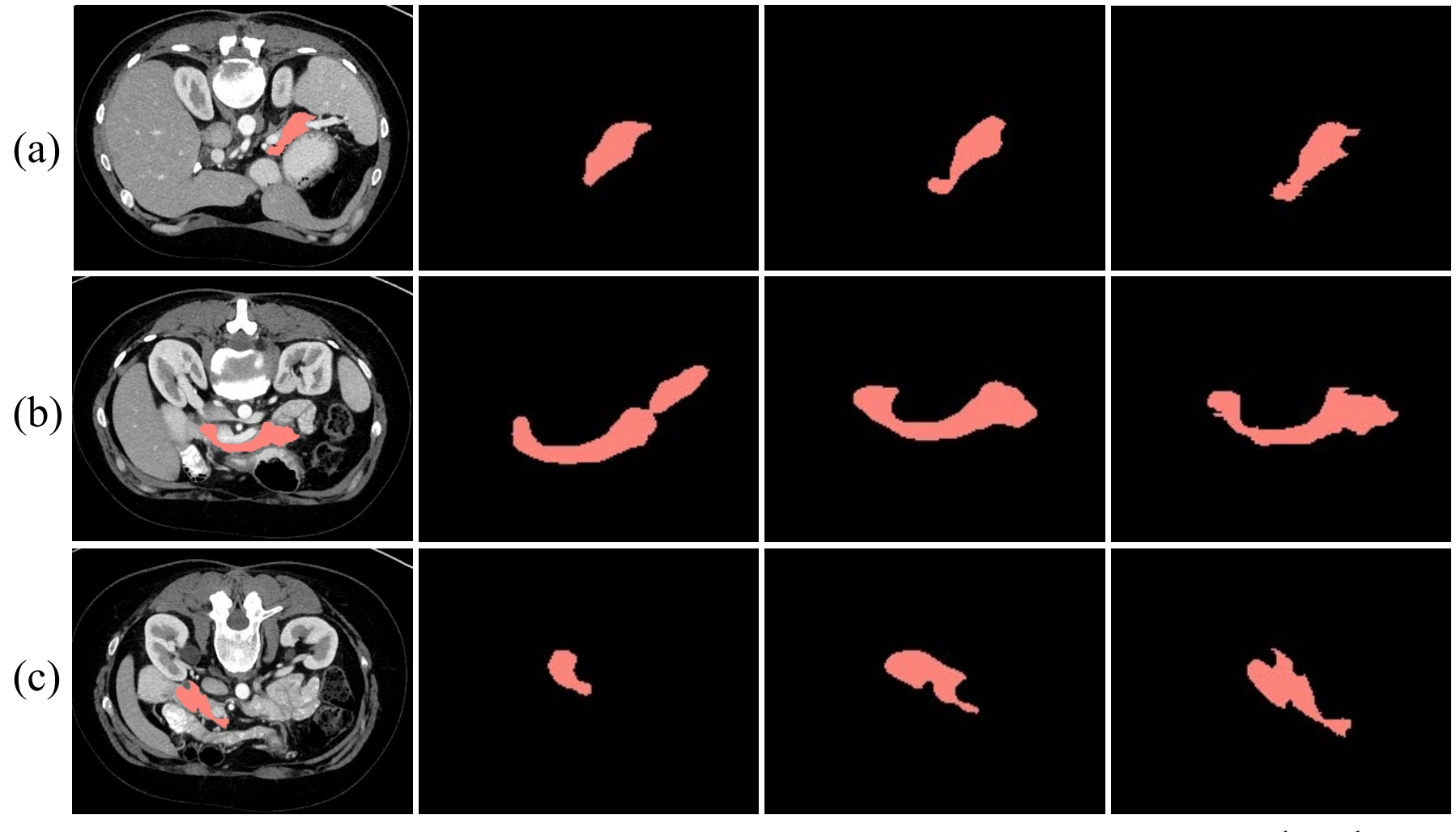

Fig. 5. Examples of pancreas segmentation results on the NIH-CT dataset. (a), (b), (c) are 2D examples of pancreatic head, middle, and tail parts, respectively. For each slice, the images from left to right are input CT slice, prediction results of U-Net and MDS-Net, and corresponding ground truth.

TABLE II. ABLATION STUDIES OF PROPOSED METHOD

| | mean ± stdv [min, max] | | |
|---|---|---|---|
| | DSC (%) | Jaccard (%) | RMSE (mm) |
| U-Net | 79.0±7.7[42.1,88.3] | 66.4±9.7[27.4,80.6] | 6.1±5.0[1.7,30.3] |
| 2.5D U-Net | 81.2±6.5[58.3,89.4] | 69.8±8.9[41.1,82.1] | 3.6±2.4[1.5,16.7] |
| **MDS-Net** | **84.5±5.2[68.3,91.2]** | **73.1±7.5[51.1,83.8]** | **3.2±1.9[1.5,9.7]** |
| **MDS-Net + SW** | **85.7±4.1[73.2,91.6]** | **75.3±6.1[57.7,84.5]** | **2.8±1.2[1.4,6.8]** |

TABLE III. COMPARISON TO OTHER STATE-OF-THE-ART MODELS ON NIH DATASET

| Index | DSC (%) | Jaccard (%) |
|---|---|---|
| Zhang *et al.* [16] | 77.9 ± 8.5[43.7, 89.2] | - |
| Roth *et al.* [5] | 78.0 ± 8.2[34.1, 88.7] | - |
| Roth *et al.* [17] | 81.3 ± 6.3[50.7, 89.0] | 68.8 ± 8.1[33.9, 80.1] |
| Zhou *et al.* [6] | 82.4 ± 5.7[62.4, 90.9] | - |
| Cai *et al.* [13] | 82.4 ± 6.7[60.0, 90.1] | 70.6 ± 9.0[42.9, 81.9] |
| Yu *et al.* [18] | 84.5 ± 5.0[62.8, 91.0] | - |
| **Ours** | **85.7 ± 4.1[73.2, 91.6]** | **75.3 ± 6.1[57.7, 84.5]** |

### D. Ablation Study

To evaluate the effect of the model-driven inter-slice regularization term and sliding window (SW) fusion algorithm, we performed ablation studies among four models: baseline U-Net, stack-based U-Net (2.5D U-Net), MDS-Net (2.5D U-Net with inter-slice regularization term), and MDS-Net with SW algorithm. As listed in Table II, the mean DSC and Jaccard accuracy of the MDS-Net is 84.5% and 73.1%, and the mean RMSE is 3.2 mm, which outperforms the benchmark U-Net and 2.5D U-Net. Moreover, the SW algorithm further improved the segmentation performance to average 85.7% DSC accuracy, 75.3% Jaccard accuracy, and 2.8 mm RMSE. These ablation experiments verify the advantage of our proposed method on pancreas segmentation.

### E. Comparison to Other State-of-the-Art Methods

We compared our proposed approach with other state-of-the-art methods in the literature [5, 6, 13, 16-18] in terms of the DSC and Jaccard indices for pancreas segmentation on the NIH-CT dataset, as shown in Table III. Overall, our MDS-Net model outperforms each of the state-of-the-art models, with an average 85.7% DSC accuracy and 75.3% Jaccard accuracy. The lower standard deviation of segmentation results can also reflect the stability and reliability of MDS-Net. Moreover, the minimum DSC accuracy of MDS-Net is 73.2%, which is much higher than other methods, demonstrating the outstanding ability of our approach for poor cases.

## IV. CONCLUSION

In this work, we presented a novel model-driven stack-based U-Net method to address the challenge of pancreas segmentation problem. Specifically, each 3D CT scan was divided into multi-stack data, and a stack-based U-Net architecture was designed for integrating local spatial context information. To ensure the segmentation accuracy of each slice in a stack, we embedded a model-driven regularization strategy into a data-driven deep learning method for constraining the inter-slice relationship. Besides, the sliding window algorithm was introduced to improve the continuity of segmentations between adjacent stacks and slices. We extensively validated the efficacy of our framework on the NIH-CT public dataset and compared our proposed method with other state-of-the-art methods. The results demonstrated that the proposed method could substantially improve the pancreas segmentation accuracy, as well as promote the stability and robustness of segmentation performance. Our proposed method provides a potential tool for the academic research and development of new clinical investigations of the pancreas.